\documentclass[runningheads,a4paper]{llncs}

\usepackage{amssymb}
\usepackage{amsmath}
\usepackage{graphicx}

\newcommand{\keywords}[1]{\par\addvspace\baselineskip
\noindent\keywordname\enspace\ignorespaces#1}

\newcommand{\gate}{\mathbf{g}}
\newcommand{\gates}{\gate^*}
\newcommand{\xcrossmap}{\mathrm{c}}

\newcommand{\boxpred}{\mathrm{b}^{pred}}
\newcommand{\boxgt}{\mathrm{b}^{gt}}

\begin{document}

\mainmatter  

\title{Depth-Adaptive Computational Policies for Efficient Visual Tracking}


%
%
\author{Chris Ying\thanks{Work done as student at the Machine Learning Department, CMU}\inst{1} \and Katerina Fragkiadaki\inst{2}}
\authorrunning{Depth-Adaptive Computational Policies for Efficient Visual Tracking}

\institute{Google Brain, 
Mountain View, CA,  
\texttt{chrisying@google.com} \and
Machine Learning Department, 
CMU, 
Pittsburgh, PA, 
\texttt{katef@cs.cmu.edu}}

%
%

\toctitle{Depth-Adaptive Computational Policies for Efficient Visual Tracking}
\maketitle

\let\thefootnote\relax\footnotetext{\it{Proceedings of the 11th International Conference on Energy Minimization Methods in Computer Vision and Pattern Recognition}, Venice, Italy, LNCS 2017. Copyright 2017 by the author(s).}

\begin{abstract}
Current convolutional neural networks algorithms for video object tracking spend the same amount of computation for each object and video frame. However, it is harder to track an object in some frames than others, due to the varying amount of clutter, scene complexity, amount of motion, and object's distinctiveness against its background. We propose a depth-adaptive convolutional Siamese network that performs video tracking adaptively at multiple neural network depths. Parametric gating functions are trained to control the depth of the convolutional feature extractor by minimizing a joint loss of computational cost and tracking error. Our network achieves accuracy comparable to the state-of-the-art on the VOT2016 benchmark. Furthermore, our adaptive depth computation achieves higher accuracy for a given computational cost than traditional fixed-structure neural networks. The presented framework extends to other tasks that use convolutional neural networks and enables trading speed for accuracy at runtime.
\end{abstract}

\section{Introduction}

Multilayer neural networks are the defacto standard machine learning tools 
for many tasks in computer vision, including visual tracking \cite{bertinetto2016fully}.
Current visual trackers  use a fixed amount of computation for every object and video frame \cite{bertinetto2016fully} \cite{ma2015hierarchical} \cite{7410714} \cite{NIPS2013_5192} \cite{Weng:2006:VOT:1223195.1223208}.  However, different video scenes have varying amount of complexity, background clutter, object motion, camera motion, or frame rate.    Fixed compute architectures  do not adapt to the difficulty of the input and are can be suboptimal computation-wise.


\begin{figure}[t]
    \centering
    \includegraphics[width=1.0\linewidth]{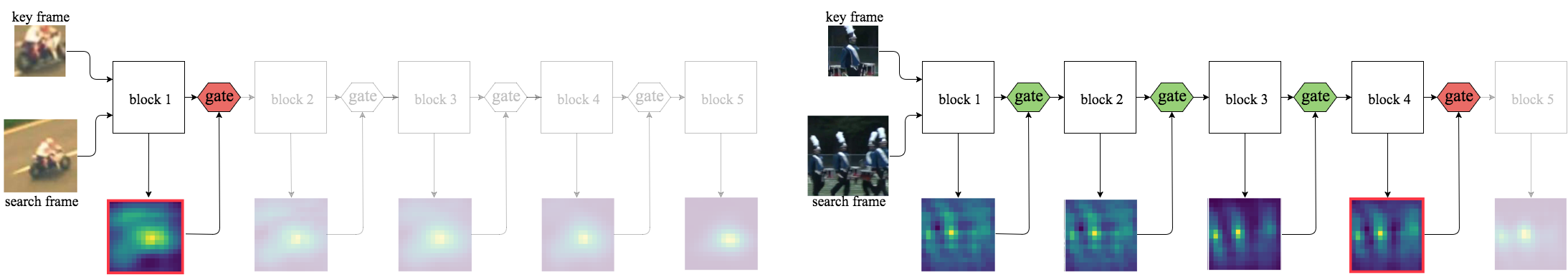}
    \centering
    \caption{\textbf{Adaptive neural computational policies for visual tracking.} 
    At each frame, we match the key frame depicting the object of interest to a search region cropped around the location of the detected object in the previous frame. 
    Our controllers (gates) decide how many blocks of the VGG  net \cite{Simonyan14c} (divided into 5 blocks of layers) to compute before computing a cross-correlation map and determining the target location.  
    In general, deeper layers yield more accurate predictions but also require more computational power. 
    Our depth-adaptive model picks the first depth for the uncluttered scene on the left and the fourth for the cluttered scene on the right.
    Red gate color denotes halting of computation at that gate.
    }
\label{fig:motivation}
\end{figure}

In this work, we propose neural architectures whose  computation adapts  to the difficulty of the task from frame to frame, rather than being fixed at runtime. 
We focus on the task of visual tracking in videos. We present  learning algorithms for training computational policies that control the  depth of Siamese convolutional networks \cite{koch2015siamese} for video tracking. 
Siamese convolutional networks track by computing cross-correlation maps of deep features, between a \textit{key frame}, where the object is labeled, and a \textit{search frame}, where the object needs to be localized, as shown in Figure \ref{fig:xcorr}. The peak of the cross-correlation map denotes the presence of the target object.  
We observe that in many search frames, the tracked object is  similar to the object in the key frame and  distinct from its background, and it would be computationally wasteful to compute  elaborate features for its detection.
To address this, we propose \textit{conditional} computation controlled by gating functions that dynamically determines how many layers of our convolutional feature extractor should be computed before computing the cross-correlation map and thus the 
target's location. 
In Figure \ref{fig:motivation}, our model uses only the first block of convolutional layers to find the motorcyclist against road, but uses 4 convolutional blocks to find the correct drummer among visually-similar peers.
Our gate controllers are trained in an end-to-end differentiable framework without the need for  sample-intense reinforcement learning. 
We test our model on the challenging VOT2016 dataset \cite{kristan2015visual} and demonstrate that we perform video tracking with close to state-of-the-art accuracy at real-time speeds.

\section{Related Works}

\subsection{Metric learning for visual tracking}

Metric learning approaches for visual tracking learn an appearance distance function between image box crops, so that the distance is large between image crops depicting different objects, and the distance is small between image crops depicting deformations of the same object instance. An accurate distance function then can be used to localize an object by computing the distances between the key frame  and various crops within the search frame. Such a distance function can be learned using a) Siamese networks \cite{koch2015siamese}, which use the same neural network weights  to extract features from a pair of images before using a single fully connected layer to predict the distance, they are trained using contrastive loss function that minimizes distance between same instance examples and requires distances to be above a certain margin for dissimilar examples. b) Triplet networks, 
\cite{DBLP:journals/corr/HofferA14} trained with a ranking loss that ensures distance of positive pairs is lower than the distance of negative pairs, and obviates the need of a margin hyper-parameter.

\subsection{Conditional computation}
Conditional computation refers to activating different network components depending on the input and serves as a promising way to reduce computational cost without sacrificing representational power. In \cite{DBLP:journals/corr/BengioBPP15},  conditional computation is implemented by selectively activating different weights in each layer and is trained via reinforcement learning. \cite{DBLP:journals/corr/ShazeerMMDLHD17} uses a sparse gating function to determine which sub-networks to execute (each of which are "experts" for different inputs),  and shows that it is possible to train the gating and network weights jointly via back-propagation. 
Graves \cite{graves2016} proposed an adaptive computation model for Recurrent Neural Networks (RNNs), that determines (depending on the input) the number of computational (pondering) steps required before producing an accurate output. 
Recent work \cite{figurnov2017}, adapts this model to convolutional networks for object detection in static images, 
where the network is trained to learn  the number of convolutional layers to be evaluated per image location,  
e.g.,``easy" image regions (e.g., sky) should require less computation than more feature-rich ones  (e.g., a car). 
Our work differs in that a) we use conditional computation in videos, rather than static images, and b) the input we predicate computational decisions on is the quality of a cross-correlation tracking map, as opposed to image classification accuracy.

\subsection{Estimating or back-propagating gradients}
A central question in all works that learn adaptive computation policies is how to train discrete \textit{gates}/\textit{controllers}, the discrete elements that determine how computation should be scheduled. Researchers have typically used non-differentiable score function estimators (a.k.a.\ REINFORCE \cite{williams1992}) for estimating the gradient with respect to such binary thresholds.  
REINFORCE has been shown to yield gradients with very high variance and requires too many samples for informative gradients to be estimated. High sample complexity is an attribute of  many other model-free RL methods, e.g., Q-learning \cite{DBLP:journals/corr/MnihKSGAWR13}. 
Indeed, recent works that use such RL techniques for neural net architectural search \cite{DBLP:journals/corr/ZophL16} or  conditional computation \cite{liu2017dynamic}, only scale to small networks and datasets \cite{liu2017dynamic},  or use large computational resources for training, e.g., in recent work 800 GPUs were used concurrently \cite{DBLP:journals/corr/ZophL16},  as opposed to a single GPU in our case.  
Alternatively, researchers have used soft, differentiable gates 
with carefully designed differentiable architectures for image generation (DRAW \cite{gregor2015_ICML}), accessing an external memory (Neural Turing Machines \cite{graves2014_neural}), deciding halting of a recurrent networks \cite{graves2016}, etc. Our training scheme, which similarly uses soft and differentiable gates during training to provide meaningful gradients, allows us to scale our policies to controlling deep neural architectures. 

\section{Depth-adaptive fully-convolutional Siamese networks}

\begin{figure}
\centering
\includegraphics[width=0.7\textwidth]{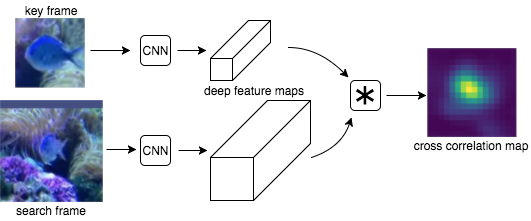}
\caption{\textbf{Siamese network with 2D cross-correlation for key-search frame pairs.} The deep feature maps for the key and search frames are extracted by the same convolutional neural network. $\ast$ denotes 2D cross-correlation.}
\label{fig:xcorr}
\end{figure}

Our model builds upon fully-convolutional Siamese  networks from \cite{bertinetto2016fully}, a state-of-the-art model for visual object tracking, which uses the same convolutional neural network to extract deep features from the key and search frames. The model then uses 2D cross-correlation to efficiently calculate the similarity score of the object in the key frame to every spatial location in the search frame, as shown in Figure \ref{fig:xcorr}. This implicitly implements a triplet network-like loss by penalizing all the negative locations and increasing the similarity at the true location. Unlike \cite{bertinetto2016fully}, which uses an AlexNet-like \cite{krizhevsky2012imagenet} architecture and trains from scratch using  ImageNet Video dataset \cite{ILSVRC15}, we use a VGG  feature extractor \cite{Simonyan14c}  pretrained from the ImageNet static image classification dataset.

We extend the fully-convolutional Siamese network for depth-adaptive computation by first dividing the convolutional layers into 5 "blocks" of convolutions and adding intermediate cross-correlations after every convolutional block. To finetune the convolutional weights, we calculate the softmax cross-entropy loss $\mathcal{L}_i$ between each of the computed cross-correlation maps $\xcrossmap_i$ and a ground-truth map $G$ for $i = 1, ..., 5$ in our tracking training set. The ground-truth map is a 2D Gaussian centered at the true location of the object in the search frame. 

\begin{equation}
\mathcal{L}_i = \texttt{softmax-cross-entropy}(\xcrossmap_i, G)
\end{equation}

\begin{figure}
\centering
\includegraphics[width=0.8\textwidth]{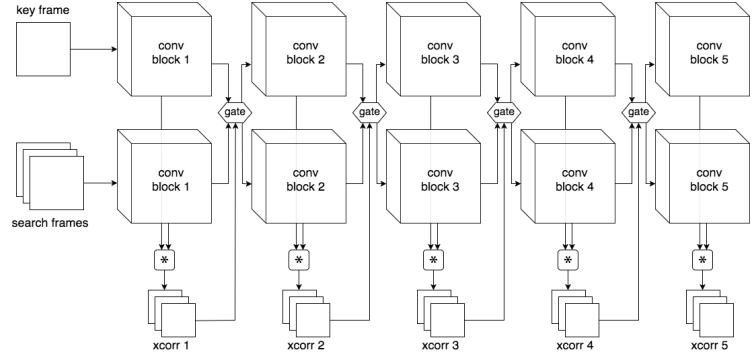}
\caption{\textbf{Depth adaptive Siamese convolutional networks.} 
Convolutional weights are shared between the two network stacks. 
Each conv block includes 2-4 convolutions with ReLU activation. 
The feature maps of the key and search frames at the end of each block are cross-correlated to yield 5 cross-correlation (\texttt{xcorr}) maps. A gating function is added at the end of each convolutional block that controls whether the network stops at this layer, or continues computation to a higher depth.}
\label{fig:arch}
\end{figure}

Furthermore, we introduce parametric gating functions between each of the convolutional blocks, which act as controllers for the depth of the VGG feature extractor at runtime.
These gating functions take as input the cross-correlation map computed using the deep features at the current depth, and output a \textit{confidence score for halting computation} at that particular depth.  In theory, we could use a convolutional neural network to extract the confidence score from each cross-correlation map. 
However, we would like our gating functions to be computationally inexpensive, so instead we use a small set of intuitive features to capture
the "quality" (certainty) of each cross-correlation map,
 such as, kurtosis (measures "peakiness"), entropy, top-5 max peak values, and the first 5 moments. Let $f$ denote the shallow feature extractor that given a cross-correlation map outputs the features above. We then learn a linear predictor, parameterized by $\phi_i$ to output the confidence score $\gate_i$ for the gate at each depth, re-scaled to $(0,1)$ via a sigmoid function, as follows: 
\begin{equation} \label{eq:gate}
\gate_i(\xcrossmap_i;\phi_i)= \mathrm{sigm}(f(\xcrossmap_i)^{T} \cdot \phi_i) \ \in \ (0,1).
\end{equation}
Our full depth-adaptive model is depicted in Figure \ref{fig:arch}. At training time, we use soft gates in order to use back-propagation for learning the gate weights, and at test time, we use hard gate thresholding, to halt computation at a particular network depth.

To train the model effectively and achieve a satisfactory trade-off of tracking accuracy and computational savings, we found the following two design choices to be crucial: 
\begin{enumerate}
\item \textbf{Intermediate supervision}: Rather than training using the loss at the deepest layer only, like \cite{bertinetto2016fully}, we  use a sum of tracking losses at all layers. This introduces intermediate supervision, which has been found to be useful in non-adaptive computational architectures such as \cite{szegedy2015going} and \cite{DBLP:journals/corr/XieT15}.
\item \textbf{Budgeted gating}: We found that directly using the confidence scores $\gate_i$ is insufficient for learning a depth-adaptive policy since each score does not affect the scores at other depths, which leads to polarized policies (either always use the shallowest depth or always use the deepest depth). Instead, we use a "budgeted" confidence score $\gates_i$, in Equation \ref{eq:budget}, where the scores sum to 1.0 and we have the desired behavior that a higher confidence score in a shallower depth corresponds to less need for deeper layers and vice-versa.

\begin{equation} \label{eq:budget}
\gates_i(\xcrossmap_i;\phi_i) = \begin{cases}
            (1 - \displaystyle \sum_{j=1}^{i-1} \gates_j(\xcrossmap_j;\phi_j))\gate_i(\xcrossmap_i;\phi_i), & i \in [1,4]\\
            1 - \displaystyle \sum_{j=1}^4 \gates_j(\xcrossmap_j;\phi_j), & i = 5.
        \end{cases}
\end{equation}
\end{enumerate}

We cannot  train the gate parameters $\phi_i, i=1 \cdots 5$ and VGG  weights jointly, since the gate feature extractor $f$ is non-differentiable. Thus, we train in two phases. In the first phase, we finetune the VGG  weights by minimizing the non-gated loss at all depths $\mathcal{L}^{\text{conv}}$:

\begin{equation}
\mathcal{L}^{\text{conv}} = \sum_{i=1}^5 \mathcal{L}_i
\label{eq:trackingloss}
\end{equation}

In the second phase, we fix the VGG  weights and train the gate parameters 
by minimizing a loss function $\mathcal{L}^{\text{gate}}$ that combines tracking loss and computational cost in all depths:
\begin{equation}
\mathcal{L}^{\text{gate}} = \underbrace{\sum_{i=1}^5 \gates_i \mathcal{L}_i}_{\text{tracking loss}} + \lambda \underbrace{\sum_{i=1}^5 p_{i} \gates_i}_{\text{computational cost}},
\end{equation}
where the hyper-parameter $\lambda$ trades off tracking accuracy and computational efficiency. We found that $\lambda \in [0.5, 1.0]$ resulted in a diverse set of depths of computation (greater or less than that range generally led to "polarized" results, either all deepest layer or all shallowest layer).
The parameter $p_i$  encodes the relative incremental computational cost of each successive layer. For our experiments, we set each $p_i$ to the incremental additional cost as reported in Table \ref{tab:flopcomp} with the $p_1=1.0$. For example, $p_2=2.43 - p_1=1.43$.

Much like \cite{DBLP:journals/corr/ShazeerMMDLHD17}, our training method provides balanced updates to all gates, meaningful gradients, and requires less training data. 
Though soft gates are used during training to enable back-propagation, at runtime, we threshold the budgeted confidence score and halt computation at a gate if the score exceeds some tune-able value. The score is a value in $[0, 1]$ and in our experiments we set the threshold at $0.25$, $0.5$, or $0.75$ for increasing degrees of strictness (i.e. higher threshold means we are less likely to accept the tracking result at a shallower layer).

\subsection{Implementation details}
The base architecture we use for the convolutional layers is the 19-layer VGG architecture \cite{Simonyan14c}. 
We remove the fully connected layers of the architecture and treat the remaining convolutional and max-pool layers as the feature extractor.
The VGG architecture is divided into 5 blocks of convolutions with 2-4 convolutional layers each, each ending in a max-pool layer.
We remove the last max-pool layer in order to keep the deep feature maps as large as possible (i.e. $16 \times 16$ in the last layer). To improve training, we normalize the key and search feature maps via batch normalization and rescale the output cross-correlation map to $[0,1]$. Cross-correlation is an expensive operation so to keep the computational costs low, we downsample the feature maps to $16 \times 16$ before cross-correlation.

Since training is performed with a single key frame and a batch of search frames, cross-correlation can be efficiently implemented on GPU by performing 2D convolution on the search feature maps with the key feature maps as the filter, treating the feature channels as the input channel size (the output channel size is 1).

Our model is implemented in TensorFlow v1.0.0 \cite{tensorflow2015-whitepaper} using pretrained VGG network weights on  ImageNet \cite{ILSVRC15} for image classification. All training and evaluation was performed on a single NVIDIA TITAN X GPU, an Intel Xeon E5-2630 v3 CPU, and 16 GB of RAM.

To efficiently implement hard-gating, we use TensorFlow's control flow operators (\verb!tf.cond!). Hard-gating is only fully efficient when the batch size of the search frames is $1$ since the computation is bottlenecked by the deepest cross-correlation map that is required by a sample in a batch. In practice, this is not as much of an issue since consecutive frames tend to use  similar depths for prediction.

\section{Experiments}

We train and test our model on the Visual Object Tracking dataset VOT2016 \cite{kristan2015visual}. The dataset consists of 60 videos with a total of 21455 frames of various resolutions. Each frame is labelled with the box corner coordinates of a bounding box that corresponds to a single object being tracked in the video. The videos have noisy backgrounds, the object can change shape or orientation, and there is occlusion in some frames. Since the VOT2016 dataset does not include a train-validation split, we randomly pick  25\% of the videos to hold out as the test set. Note that the VOT dataset is designed for an evaluation-only competition so our results are not directly comparable to existing benchmarks. Our goal is not necessarily to beat state-of-the-art methods, but rather to present a useful technique for fast video tracking which can improve nearly any convolutional model.

We preprocess the videos by selecting a key frame every $10$ frames and the subsequent up-to-$100$ frames as the search frames. We resize and crop the key frames to $128 \times 128$ centered at the tracked object such that there is at least $25\%$ padding around the bounding box. Each of the search frames are resized with the same scale and cropped to $256 \times 256$ such that the frame is centered at the object at the previous frame. If the cropped search frame extends beyond the edge of the image, we pad the extra pixels with the mean RGB value of the dataset. The predicted object box is found using the position of the maximum value in the cross-correlation map as the offset and the bounding box dimensions are the same as the key frame reference box.

\subsection{Evaluation metrics}

We measure tracking accuracy using Intersection-Over-Union (IOU) ] between the predicted object box $\boxpred$ and the ground truth box $\boxgt$:
\begin{equation} \label{eq:iou}
    \text{IOU}(\boxpred, \boxgt) = \frac{|\boxpred \cap \boxgt|}{|\boxpred \cup \boxgt|}.
\end{equation} 
We measure  IOU at up-to 1, 5, and 25 frames ahead of the key frame, e.g., for IOU@25, we take the average IOU of the tracker with key frame $t$ and search frames $t+1, t+2, ..., t+25$.  
The larger the frame gap between key frame and search frame, the more the tracking target deforms and the harder it is to track.  

We measure computational cost by computing the number of floating point operations (FLOPs) required to perform tracking on a batch of $25$ search frames. Experimentally, we find that FLOPs is a good proxy for true computational cost as measured in frames-per-second (FPS). The reason FLOPs is the preferable metric is that FPS is heavily tied to hardware and software constraints, which may prevent the architecture from achieving the theoretical speedup.

\subsection{Siamese tracker performance}

\begin{figure}
\centering
\includegraphics[width=1.0\textwidth]{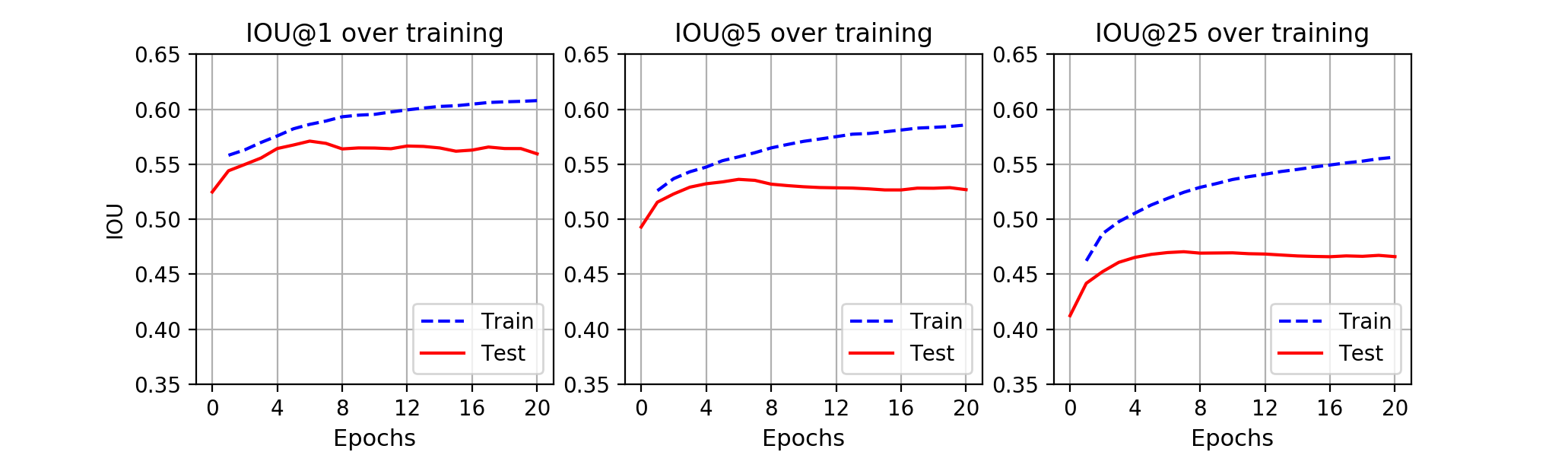}
\caption{\textbf{Finetuning.} Training and testing IOU curves during  metric learning  for different frame gaps. Starting with weights pretrained on ImageNet image classification, our VGG feature extractor fast reaches top performance.
}
\label{fig:finetune1}
\end{figure}

We finetune our VGG feature extractor starting from pretrained weights using the tracking loss of Eq. \ref{eq:trackingloss}. The tracking performance during finetuning is shown in Figure \ref{fig:finetune1}. Using pretrained weights allows the model to reach peak performance after only a few epochs. For this evaluation we use the full network depth.

\begin{figure}
\centering
\includegraphics[width=0.5\textwidth]{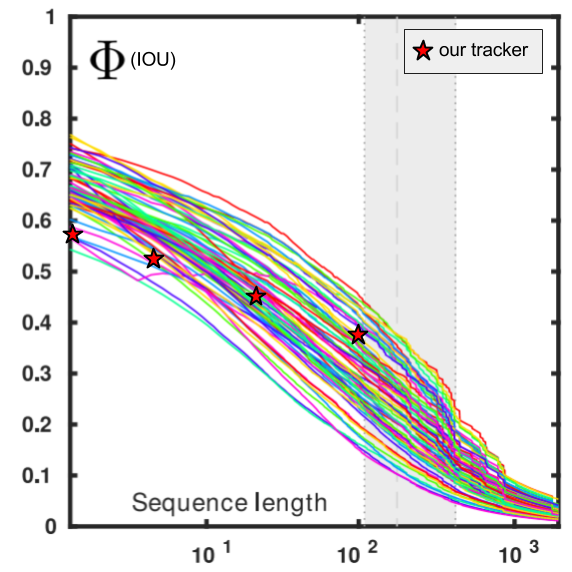}
\caption{\textbf{VOT2016} Comparison of our full-computation model against top submissions to VOT2016 \cite{Kristan2016a}. The stars represent our tracker's accuracy at selected sequence lengths.}
\label{fig:finetune2}
\end{figure}

Our implementation is comparable to the top submissions to the VOT2016 competition \cite{Kristan2016a} in IOU over sequence length, as seen in Figure \ref{fig:finetune2}.
Though it is not as accurate as the best trackers at small sequence lengths, it is competitive with many trackers at around 100 frames. Note also that the top four trackers from VOT2016 run at under 1 FPS while our system runs at over 54 FPS using the shallowest layer (\texttt{xcorr1}) and over 37 FPS using the deepest layer (\texttt{xcorr5}). Note that our code was not optimized so FPS is not a very good metric for computational cost. Additional software engineering (e.g. multithreaded inputs, better GPU utilization) should bring the \texttt{xcorr1} FPS to well over 100.

\subsection{Effect of intermediate supervision} \label{sec:intermediate}
\begin{figure}
\centering
\includegraphics[width=0.8\textwidth]{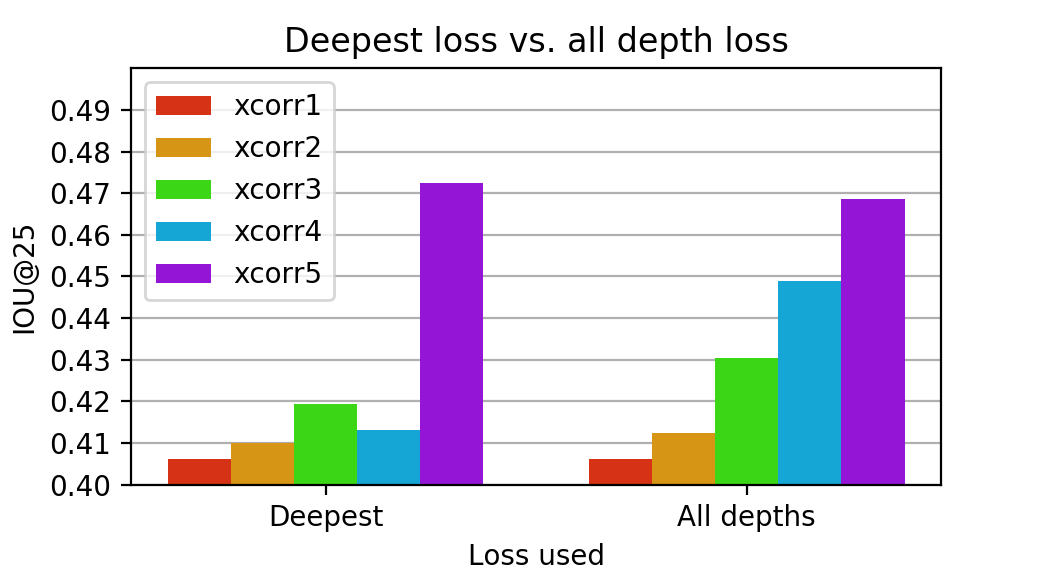}
\caption{\textbf{Intermediate supervision.} Supervision at intermediate layers (as opposed to the top layer only)  increases the accuracy of the intermediate  cross-correlation maps, and makes depth-adaptation at runtime worthwhile.}
\label{fig:finetunecomp}
\end{figure}

We compare the performance of the tracker when trained with loss only at the deepest layer against the tracker when trained with losses added in all depths. The IOU@25 comparison is shown in Figure \ref{fig:finetunecomp}.

Training with all intermediate losses yields increased accuracy as depth increases,  while training with only the deepest yields a big jump in accuracy at depth 5 but lower accuracy at shallower depths. If computational cost is not a factor, using only the deepest loss gives around 0.01 IOU benefit over using all depths. Since our goal is to use depth-adaptive feature extraction, intermediate supervision is essential. 

\subsection{Depth-adaptive computational policies}

The computational policies we compare are:
\begin{itemize}
\item Fixed-depth: always use the cross-correlation map at a fixed depth (i.e. \texttt{xcorr1}, ... \texttt{xcorr5}).
\item Soft-gating: use a sum of the cross-correlation maps at all the depths, weighted by the budgeted confidence score. This policy does not save computational cost but serves as a baseline that also utilizes the gating functions.
\item Hard-gating: we halt computation if the budgeted confidence score exceeds a tune-able threshold, which is a model hyper-parameter. 
We report  performance while varying hyper-parameters in order to obtain accuracy/computation trade offs across the whole spectrum. 
\end{itemize}

Table \ref{tab:flopcomp} shows the theoretical FLOPs required to compute the cross-correlation maps for a single key-search batch of 1 key frame and 25 search frames. For simplicity of calculation, the values only include the floating point multiplication operations, which comprise the overwhelming majority of the computation. As mentioned earlier, FLOPs is a better metric than FPS because it is independent of  implementation details of the algorithm. 

\begin{table}[]
    \centering
    \begin{tabular}[b]{|l|r|r|}
        \hline
        Gating policy & FLOPs ($\times 10^9$) & Relative to \texttt{xcorr1} \\
        \hline
        \texttt{xcorr1} & $2.78$ & $1.00\times$ \\
        \texttt{xcorr2} & $67.70$ & $2.43\times$ \\
        \texttt{xcorr3} & $160.75$ & $5.78\times$ \\
        \texttt{xcorr4} & $253.79$ & $9.12\times$ \\
        \texttt{xcorr5} & $280.37$ & $10.07\times$ \\
        soft-gating & $280.53$ & $10.08\times$ \\
        hard-gating & varies & varies \\
        \hline
    \end{tabular}
    \vspace{0.3cm}
    \caption{\textbf{Theoretical FLOPs for varying  network depth.} \texttt{xcorr}$i$ denotes network evaluation up until the $i$-th cross-correlation  map. Soft-gating uses all five cross-correlation maps weighted by the gating confidence. The gating feature computation is negligible in comparison to convolutional feature extraction.}
    \label{tab:flopcomp}
\end{table}

\begin{figure}
    \centering
    \includegraphics[width=0.8\textwidth]{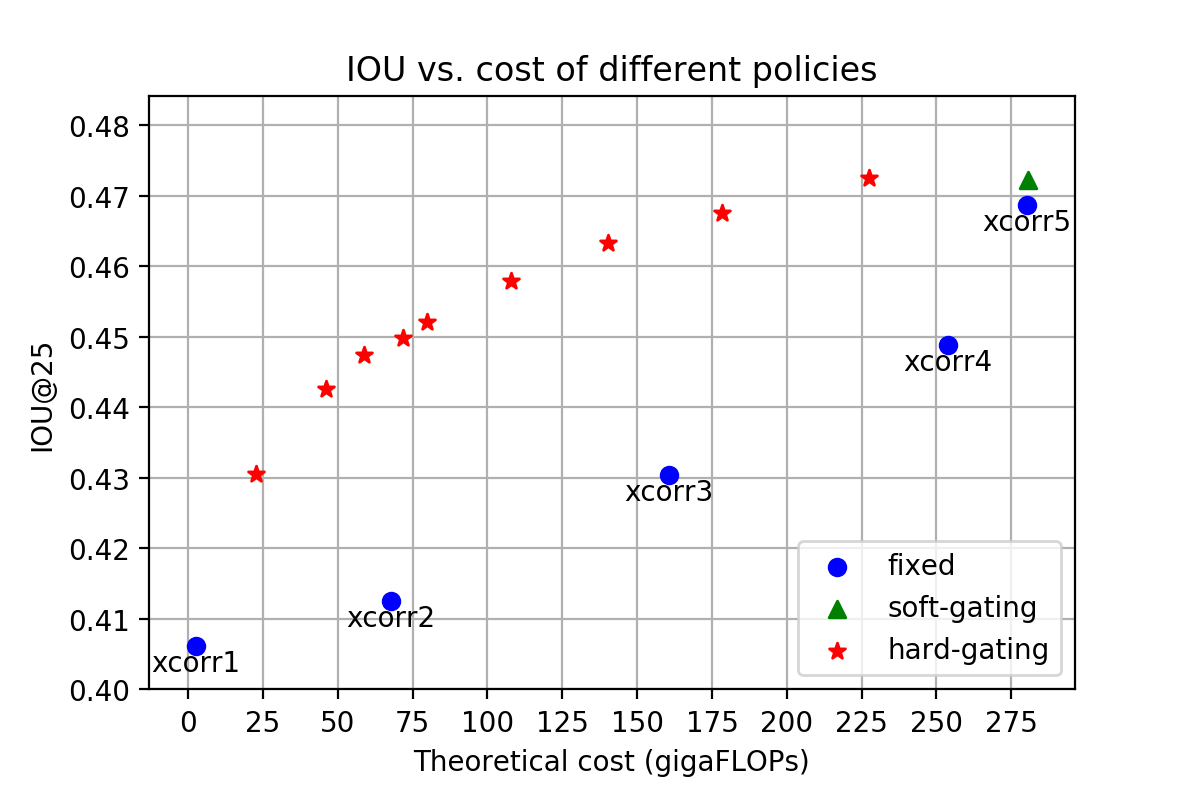}
    \caption{\textbf{Accuracy versus  computation curves} for our model and baselines. We generated the curve for our model (\textit{hard gating}) by varying the relative weight $\lambda$ of computational cost and tracking accuracy. For our fixed depth baseline, we obtain five points by varying the number of convolutional blocks from 1 to 5. Top left of the diagram is more desirable. Our model clearly outperforms the non-learned fixed-depth policies.}
    \label{fig:policies}
\end{figure}

Figure \ref{fig:policies} compares the accuracy and computational cost of each of the policies. Both soft and hard-gating can achieve accuracy values that exceed any fixed-depth policy and furthermore hard-gating uses significantly less computational cost to achieve the same or better accuracy. By varying hyper-parameter  $\lambda$, we achieve different trade-offs of accuracy and computational cost depending on the requirements of the task.

\begin{figure}
\centering
\includegraphics[width=1.0\textwidth]{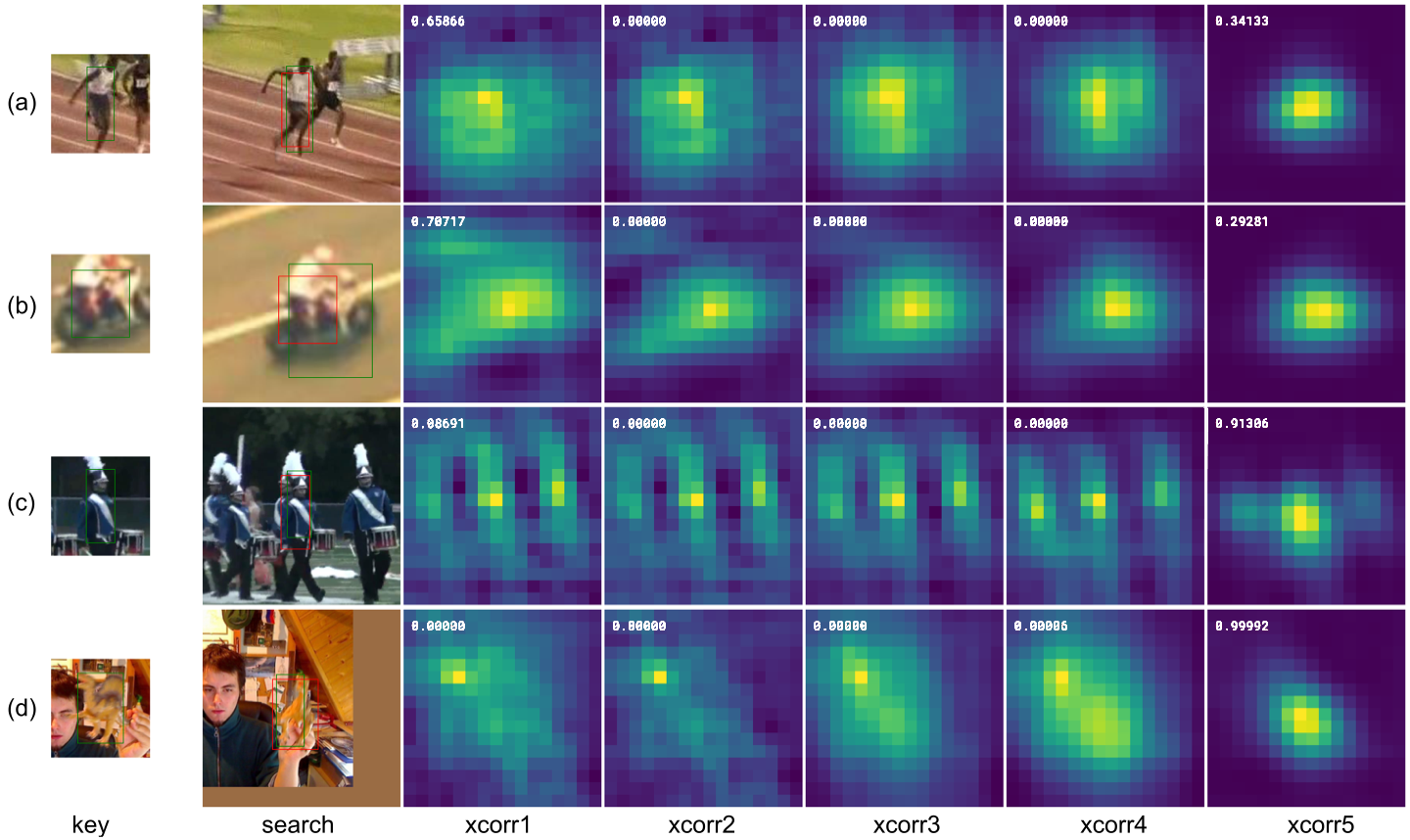}
\caption{\textbf{Tracking results.} Green box is ground truth, red box is prediction, red numbers are confidence weights. In (a), the tracker learns that \texttt{xcorr1} is sufficient for tracking. In (b),  the tracker learns that it needs to compute \texttt{xcorr5} in order to confidently track the object.}
\label{fig:output}
\end{figure}

The cross-correlation maps for selected video frames can be viewed in Figure \ref{fig:output}.

\section{Discussion}
Our experimental results show that our proposed depth-adaptive fully convolutional siamese network  successfully tracks at accurcies comparable to state-of-the-art submissions to VOT2016. We demonstrate that our learned depth-adaptive policies can outperform fixed-depth networks while using significantly less computational power. Furthermore, we show that we can easily trade accuracy for computational cost as necessary by changing the model hyper-parameters.

Our work is limited by the following factors:
\begin{itemize}
\item Our model is finetuned purely on VOT2016 data unlike most other submissions which use ImageNet VID,  or other tracking datasets for training the Siamese network. VOT2016 dataset is considerably smaller than ImageNet VID and the videos are considered more difficult to track.
\item Fully-convolutional Siamese networks do not support updating the tracking model (i.e. key frame object appearance) since it does not naively support bounding box rescaling. 
\item Our final cross-correlation map resolution is $16 \times 16$, which is too coarse for fine-grained tracking. We experimented with higher resolution cross-correlation maps  and obtained  15\% improvement to IOU. However, this method is incompatible with hard-gating since it requires cross-correlation at multiple depths.
\end{itemize}

\section{Conclusion}

We have presented  a conditional computation model for visual tracking, where computational depth is allocated based on a frame's tracking difficulty. Our  model  is comprised of continuous weight filter variables and discrete learned controllers that dynamically manage the depth of the network at runtime, and balance accuracy with  computational cost.
The proposed model saves computation on "easy" frames without sacrificing  representational power on difficult frames that require deeper features to track. We show our model outperforms naive non-adaptive policies by a significant margin, as measured by accuracy at a various computational costs. 
Paths for future work include multi-scale tracking and more complex gating features, potentially using the history of the cross-correlation maps to determine computation. Though this work investigates policies that control the depth of the network,  other promising ``actions" for conditional computation in visual tracking include  adaptive spatial computation e.g., by using motion to focus attention to the moving parts of the scene, or frame skipping e.g., by using only a subset of frames to localize the target without sacrificing tracking accuracy. 

The methods  presented in this work can be extended to any task which uses deep neural networks. By treating neural networks as a series of composable feature extractors, we have the ability to select feature embeddings at various degrees of complexity, which can both reduce computational cost and potentially improve performance. 

\bibliographystyle{abbrv}
\bibliography{bibliography}

\begin{thebibliography}{10}

\bibitem{tensorflow2015-whitepaper}
M.~Abadi, A.~Agarwal, P.~Barham, E.~Brevdo, Z.~Chen, C.~Citro, G.~S. Corrado,
  A.~Davis, J.~Dean, M.~Devin, S.~Ghemawat, I.~Goodfellow, A.~Harp, G.~Irving,
  M.~Isard, Y.~Jia, R.~Jozefowicz, L.~Kaiser, M.~Kudlur, J.~Levenberg,
  D.~Man\'{e}, R.~Monga, S.~Moore, D.~Murray, C.~Olah, M.~Schuster, J.~Shlens,
  B.~Steiner, I.~Sutskever, K.~Talwar, P.~Tucker, V.~Vanhoucke, V.~Vasudevan,
  F.~Vi\'{e}gas, O.~Vinyals, P.~Warden, M.~Wattenberg, M.~Wicke, Y.~Yu, and
  X.~Zheng.
\newblock {TensorFlow}: Large-scale machine learning on heterogeneous systems,
  2015.
\newblock Software available from tensorflow.org.

\bibitem{DBLP:journals/corr/BengioBPP15}
E.~Bengio, P.~Bacon, J.~Pineau, and D.~Precup.
\newblock Conditional computation in neural networks for faster models.
\newblock {\em CoRR}, abs/1511.06297, 2015.

\bibitem{bertinetto2016fully}
L.~Bertinetto, J.~Valmadre, J.~F. Henriques, A.~Vedaldi, and P.~H. Torr.
\newblock Fully-convolutional siamese networks for object tracking.
\newblock In {\em European Conference on Computer Vision}, pages 850--865.
  Springer, 2016.

\bibitem{figurnov2017}
M.~Figurnov, M.~D. Collins, Y.~Zhu, L.~Zhang, J.~Huang, D.~P. Vetrov, and
  R.~Salakhutdinov.
\newblock Spatially adaptive computation time for residual networks.
\newblock In {\em CVPR}, 2017.

\bibitem{graves2016}
A.~Graves.
\newblock Adaptive computation time for recurrent neural networks.
\newblock {\em CoRR}, abs/1603.08983, 2016.

\bibitem{graves2014_neural}
A.~Graves, G.~Wayne, and I.~Danihelka.
\newblock Neural turing machines.
\newblock {\em CoRR}, abs/1410.5401, 2014.

\bibitem{gregor2015_ICML}
K.~Gregor, I.~Danihelka, A.~Graves, D.~J. Rezende, and D.~Wierstra.
\newblock {DRAW:} {A} recurrent neural network for image generation.
\newblock In {\em ICML}, pages 1462--1471, 2015.

\bibitem{DBLP:journals/corr/HofferA14}
E.~Hoffer and N.~Ailon.
\newblock Deep metric learning using triplet network.
\newblock {\em CoRR}, abs/1412.6622, 2014.

\bibitem{koch2015siamese}
G.~Koch.
\newblock {\em Siamese neural networks for one-shot image recognition}.
\newblock PhD thesis, University of Toronto, 2015.

\bibitem{Kristan2016a}
M.~Kristan, A.~Leonardis, J.~Matas, M.~Felsberg, R.~Pflugfelder,
  L.~\v{C}ehovin, T.~Vojir, G.~H\"{a}ger, A.~Luke\v{z}i\v{c}, and G.~Fernandez.
\newblock The visual object tracking vot2016 challenge results.
\newblock Springer, Oct 2016.

\bibitem{kristan2015visual}
M.~Kristan, J.~Matas, A.~Leonardis, M.~Felsberg, L.~Cehovin, G.~Fernandez,
  T.~Vojir, G.~Hager, G.~Nebehay, and R.~Pflugfelder.
\newblock The visual object tracking {VOT2015} challenge results.
\newblock {\em ICCV}, pages 1--23, 2015.

\bibitem{krizhevsky2012imagenet}
A.~Krizhevsky, I.~Sutskever, and G.~E. Hinton.
\newblock Imagenet classification with deep convolutional neural networks.
\newblock In {\em Advances in neural information processing systems}, pages
  1097--1105, 2012.

\bibitem{liu2017dynamic}
L.~Liu and J.~Deng.
\newblock Dynamic deep neural networks: Optimizing accuracy-efficiency
  trade-offs by selective execution.
\newblock {\em arxiv}, abs/1701.00299, 2017.

\bibitem{ma2015hierarchical}
C.~Ma, J.-B. Huang, X.~Yang, and M.-H. Yang.
\newblock Hierarchical convolutional features for visual tracking.
\newblock {\em ICCV}, pages 3074--3082, 2015.

\bibitem{DBLP:journals/corr/MnihKSGAWR13}
V.~Mnih, K.~Kavukcuoglu, D.~Silver, A.~Graves, I.~Antonoglou, D.~Wierstra, and
  M.~A. Riedmiller.
\newblock Playing {A}tari with deep reinforcement learning.
\newblock {\em arxiv}, abs/1312.5602, 2013.

\bibitem{ILSVRC15}
O.~Russakovsky, J.~Deng, H.~Su, J.~Krause, S.~Satheesh, S.~Ma, Z.~Huang,
  A.~Karpathy, A.~Khosla, M.~Bernstein, A.~C. Berg, and L.~Fei-Fei.
\newblock {ImageNet Large Scale Visual Recognition Challenge}.
\newblock {\em IJCV}, 115(3):211--252, 2015.

\bibitem{DBLP:journals/corr/ShazeerMMDLHD17}
N.~Shazeer, A.~Mirhoseini, K.~Maziarz, A.~Davis, Q.~V. Le, G.~E. Hinton, and
  J.~Dean.
\newblock Outrageously large neural networks: The sparsely-gated
  mixture-of-experts layer.
\newblock {\em CoRR}, abs/1701.06538, 2017.

\bibitem{Simonyan14c}
K.~Simonyan and A.~Zisserman.
\newblock Very deep convolutional networks for large-scale image recognition.
\newblock {\em CoRR}, abs/1409.1556, 2014.

\bibitem{szegedy2015going}
C.~Szegedy, W.~Liu, Y.~Jia, P.~Sermanet, S.~Reed, D.~Anguelov, D.~Erhan,
  V.~Vanhoucke, and A.~Rabinovich.
\newblock Going deeper with convolutions.
\newblock In {\em CVPR}, 2015.

\bibitem{7410714}
L.~Wang, W.~Ouyang, X.~Wang, and H.~Lu.
\newblock Visual tracking with fully convolutional networks.
\newblock In {\em 2015 IEEE International Conference on Computer Vision
  (ICCV)}, pages 3119--3127, Dec 2015.

\bibitem{NIPS2013_5192}
N.~Wang and D.~yan Yeung.
\newblock Learning a deep compact image representation for visual tracking.
\newblock In C.~Burges, L.~Bottou, M.~Welling, Z.~Ghahramani, and
  K.~Weinberger, editors, {\em Advances in Neural Information Processing
  Systems 26}, pages 809--817. 2013.

\bibitem{Weng:2006:VOT:1223195.1223208}
S.-K. Weng, C.-M. Kuo, and S.-K. Tu.
\newblock Video object tracking using adaptive kalman filter.
\newblock {\em J. Vis. Comun. Image Represent.}, 17(6):1190--1208, 2006.

\bibitem{williams1992}
R.~J. Williams.
\newblock Simple statistical gradient-following algorithms for connectionist
  reinforcement learning.
\newblock {\em Machine Learning}, 8(3):229--256, 1992.

\bibitem{DBLP:journals/corr/XieT15}
S.~Xie and Z.~Tu.
\newblock Holistically-nested edge detection.
\newblock {\em CoRR}, abs/1504.06375, 2015.

\bibitem{DBLP:journals/corr/ZophL16}
B.~Zoph and Q.~V. Le.
\newblock Neural architecture search with reinforcement learning.
\newblock {\em arxiv}, abs/1611.01578, 2016.

\end{thebibliography}

\end{document}